# State-Space Abstraction for Anytime Evaluation of Probabilistic Networks

**Michael P. Wellman and Chao-Lin Liu**

University of Michigan
Artificial Intelligence Laboratory
1101 Beal Avenue
Ann Arbor, MI 48109-2110 USA
{wellman, chaolin} @engin.umich.edu

## Abstract

One important factor determining the computational complexity of evaluating a probabilistic network is the cardinality of the state spaces of the nodes. By varying the granularity of the state spaces, one can trade off accuracy in the result for computational efficiency. We present an anytime procedure for approximate evaluation of probabilistic networks based on this idea. On application to some simple networks, the procedure exhibits a smooth improvement in approximation quality as computation time increases. This suggests that state-space abstraction is one more useful control parameter for designing real-time probabilistic reasoners.

## 1. INTRODUCTION

Despite the increasing popularity of probabilistic networks (also known as belief networks, Bayesian networks, etc.) for representing and reasoning about uncertain situations, the complexity of inference in this formalism remains a significant concern. It is well known that the problem of computing a conditional probability in a probabilistic network is NP-hard (Cooper 1990), and moreover, experience has shown that intractability is the typical—not just the worst—case. Researchers in recent years have adopted a variety of strategies, including approximation (although it has recently been shown that even approximating a conditional probability to a fixed degree of accuracy is NP-hard (Dagum and Luby 1993)), restricting network structure, and employing heuristic methods to improve average performance.

The problem might be finessed when networks are handcrafted by knowledge engineers to serve as the core of a consultation system. In this case, speed vs. accuracy tradeoffs can be resolved at design time, as the modeler tunes the network to fit the deployment technology and performance requirements at hand. Increasingly, however, probabilistic networks are constructed not by patient craftspeople, but rather are automatically generated from some underlying knowledge representation (Breese et al. 1994). This includes temporal probabilistic networks, where the structure of a network fragment is replicated for each time point (Nicholson and Brady 1994; Provan 1994), as well as knowledge-based model construction systems, where the fine-grained selection of variables and relationships is customized for particular problems and circumstances (Breese 1992; Goldman and Charniak 1993; Saffiotti and Umkehrer 1994; Wellman et al. 1992).

For example, in our current work we are exploring the application of probabilistic networks to traffic applications, including the tasks of route guidance and travel-time prediction. The complete system would include general knowledge about traffic flow on roads of varying capacities and layouts, as well as specific knowledge about a particular road network and traffic patterns. To support a specific traffic management task, the system would generate a special-purpose probabilistic network geared to the spatial and temporal scope of the problem, and level of detail necessary. The overall task is real-time, as the traffic is moving while the system processes the query. Thus, the complexity of inference in the generated network is critical, as the value of a result degrades significantly as time passes. In the extreme case, if the time needed to compute a recommended route is longer than the time to travel an obvious route, then the route guidance system would be worthless.

One approach that has been suggested for dealing with real-time inference is to perform explicit decision-theoretic metareasoning and construct an optimal probabilistic network with computation time taken into account (Breese and Horvitz 1991). While this approach is ideal in principle, practical application requires specification of the effect of modeling choices on computation time, and imposes a non-negligible overhead of the metalevel optimization.

A somewhat less flexible, but more common, approach to real-time inference in probabilistic networks is to arrange the reasoning process so that it produces progressively more useful results as more computation time is allocated. In these so-called *anytime* algorithms, the value of the result might be measured in degree of approximation or tightness of bounds on the query. For example, typical stochastic simulation algorithms are anytime in that the estimate is expected to improve as the sample size increases. Others have proposed methods that bound a conditional probability by progressively accounting for more of the event instances (D'Ambrosio 1993; Horvitz et al. 1989). An advantage of the anytime approach is that we can produce inferential behavior that is sensitive to problem-specific time stress, without necessarily performing a metalevel optimization. However, there is a large space of anytime strategies, with widely varying profiles of computational value over time. Thus, we cannot really get away without some sort of metalevel evaluation, even if it is only an off-line design-time analysis.

In this paper, we explore another variety of anytime algorithm for evaluating probabilistic networks. Specifically, we consider the possibility of modulating precision in the state space of variables in order to generate results of progressively improving approximation as computation time advances. This approach is motivated by the observation that state-space cardinality can have a large impact on computational requirements (Ezawa 1986), and yet the choice of state space is often rather arbitrary. For example, in our traffic applications, many of the variables range over real intervals. The ideal partition of the interval into subintervals depends on our precision requirements and computational resources for a given problem. Rather than fixing a granularity at design time, we seek methods that can flexibly reason at varying levels of precision depending on the particulars of our problem instance.

In the remainder of this paper we describe an anytime algorithm that produces progressively improving estimates of a probabilistic query via a series of incremental refinements of the state space of random variables. We present the basic concepts, some experimental results, and a discussion of the lessons and limitations of this study.

## 2. BACKGROUND AND EXAMPLE

A probabilistic network is a directed acyclic graph $G=(V,E)$, where $V$ is a collection of nodes or state variables, and $E$ is a set of directed links encoding dependency relationships between state variables. In addition, associated with each node is a table describing the conditional probability distribution of that state variable given all combinations of values of its predecessors. For more information on probabilistic networks in general, see for example, (Charniak 1991; Neapolitan 1990; Pearl 1988).

Some of the experiments we describe in this paper employ the following example, which models a highly simplified commuting problem. In this model depicted below, the top row of nodes represent the time of day that the commuter leaves home (*LH*), arrives at work (*AW*), finishes work (*FW*), and arrives at home (*AH*), respectively. The next row of nodes represent, respectively, the time durations spent by the commuter going to work (*GW*), performing his or her work load (*WL*), and going home (*GH*). Above each node in brackets is the real interval describing the possible values of the state variable. For example, the commuter will leave home sometime between 6 and 8 AM ($LH \in [6,8]$), and will spend 7 to 8 hours at work ($WL \in [7,8]$; this is a highly idealized model). The travel times to work and home depend probabilistically on the time of departure (e.g., due to fluctuating travel patterns), and the time at work may also depend on the time of arrival. The nodes *AW*, *FW*, and *AH* are deterministic functions (simply the sum) of their predecessors, for example, $AW = LH + GW$.[1]

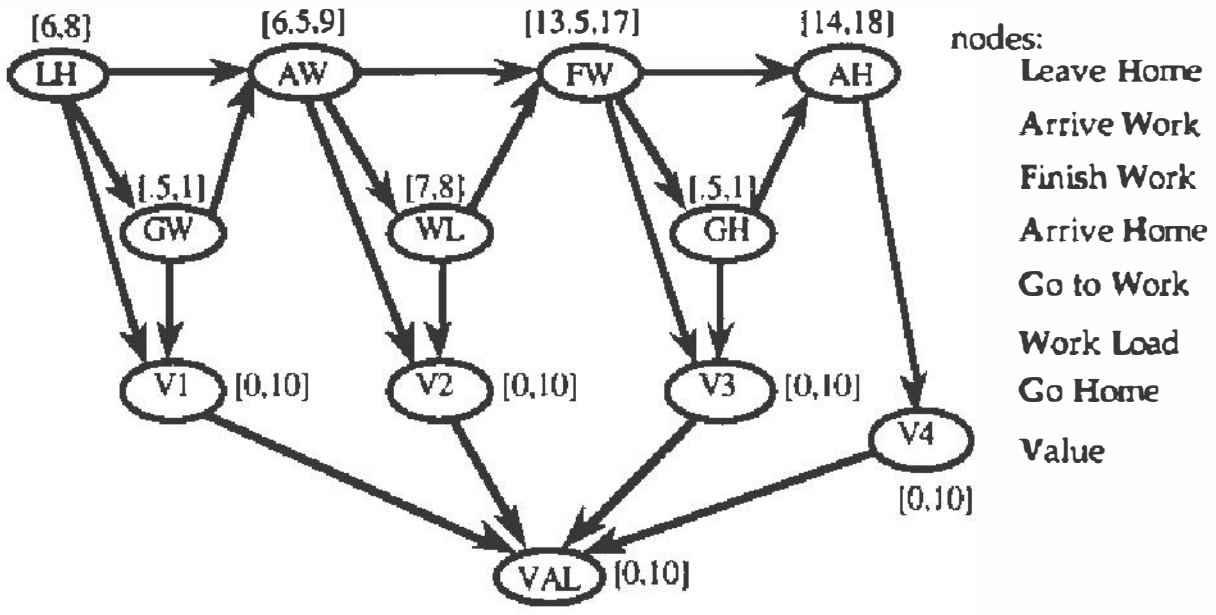

**Figure 1**: Example probabilistic network for a simplified commuter problem.

The node *VAL* represents the overall value the commuter attributes to the day's itinerary. Nodes $V1, \ldots, V4$ represent subvalue nodes, representing the value attributable to specific parts of the day. Although the intent of these nodes is to represent preferences, we treat them here as ordinary chance nodes in the probabilistic network.

Even for the simple network in Figure 1, evaluation can be complex when the state spaces of the nodes become large. Suppose we are interested in the distribution of *VAL* for a particular value of *LH*. That is, we want to evaluate a particular choice for when to leave home. Table 1 presents the CPU time[2] required for this query as a function of the number of states for each node.

| Number of states | 2 | 4 | 6 | 8 |
|---|---|---|---|---|
| CPU time (non-gc, secs.) | 1.23 | 47.0 | 484 | 2600 |

**Table 1**: Average CPU time for evaluation as a function of state cardinality.

The entry for $k$ states corresponds to the network where each node's state space is divided into $k$ equal intervals.

[1]The evaluation algorithms we employ do not exploit this deterministic relationship or its simple additive form; we have chosen it purely for simplicity of specification. In general, we have not optimized the performance of our basic evaluation methods. However, this should not affect our qualitative conclusions about the *relative* performance of the inference procedure at varying levels of abstraction.

[2]Using Jensen's method (Jensen et al. 1990) implemented in IDEAL (Srinivas and Breese 1990), running on Allegro Common Lisp on a Sun SparcStation IPX. Note that the data reported in the other experiments below were collected under different configurations.

Cases for $k > 8$ were beyond reasonable waiting for this network given the computational setup used.

# 3. STATE-SPACE ABSTRACTION

## 3.1. OVERVIEW

The idea of approximate evaluation methods is to trade accuracy for computational tractability. The statistics in Table 1 suggest that one important dimension on which to weigh this tradeoff is granularity of the state space. Coarsening a model by ignoring some distinctions among states can dramatically improve the computation time, at the expense of fidelity of the results.

The algorithm we employ begins with the coarsest possible model, in which each node has only two states. (Collapsing these two states into one is tantamount to ignoring the node, which is another option that might be considered in some circumstances.) If time is available after evaluating this model, we then refine the state space of each node by introducing another distinction. This process iterates until we either run out of time or solve the finest-grained network model. The algorithm is anytime, as the expected accuracy increases as the nodes are progressively refined.

## 3.2. ABSTRACTING A PROBABILISTIC NETWORK

In describing the abstraction procedure, the following definitions are useful.

**Definition 1** The *original probabilistic network* (OPN) is the given, finest-grain probabilistic network.

**Definition 2** An *elementary state* of a state variable is a state of that state variable in the OPN.

We assume that the elementary states are ordered, $s_1 < s_2 < \cdots < s_m$.

**Definition 3** A *superstate* of a state variable is a state that consists of two or more elementary states, adjacent in the ordering. We use the notation $[s_{i,j}]$ to refer to the superstate consisting of elementary states $[s_i, s_{i+1}, \ldots, s_j]$.

**Definition 4** An *abstract probabilistic network* (APN) is a probabilistic network in which one or more state variables have superstates.

In an abstract version of a probabilistic network, each variable is associated with a set of superstates partitioning its elementary states. The joint probability distribution with respect to the superstates can be described in terms of the elementary states making up those superstates:

$$\Pr\left(\left[s^1_{i^1,j^1}\right],\left[s^2_{i^2,j^2}\right],\ldots,\left[s^n_{i^n,j^n}\right]\right) = \sum_{k_n=i^n}^{j^n} \cdots \sum_{k_2=i^2}^{j^2} \sum_{k_1=i^1}^{j^1} \Pr\left(s^1_{k_1}, s^2_{k_2}, \ldots, s^n_{k_n}\right). \tag{1}$$

The joint distribution over elementary states is factored in the OPN into conditional probabilities:

$$\Pr\left(s^1_{k_1}, s^2_{k_2}, \ldots, s^n_{k_n}\right) = \prod_{l=1}^{n} \Pr\left(s^l_{k_l} \middle| pred(s^l)\right). \tag{2}$$

where $pred(s^l)$ denotes the values of the predecessors of $s^l$ in the OPN. We can express conditional probabilities over superstates in terms of the elementary states as well,

$$\Pr\left(\left[s^l_{i^l,j^l}\right] \middle| pred(s^l)\right) = \frac{\sum_{k=i^l}^{j^l} \sum_{s' \in pred(s^l)} \Pr\left(s^l_k \middle| s'\right)\Pr(s')}{\sum_{s' \in pred(s^l)} \Pr(s')}. \tag{3}$$

where $s'$ ranges over the elementary states constituting the predecessor values of $s^l$. Note, however, that the APN might not be factorable according to the same dependence structure as was the OPN. For example, consider the simple network

$$a \rightarrow b \rightarrow c.$$

Suppose that the elementary state space of $b$ is $[b_1, b_2, b_3]$. The network structure above is valid as long as $a$ and $c$ are conditionally independent given each of these values for $b$. However, even when this is the case, it is quite possible that $a$ and $c$ be *de*pendent given the superstate $[b_{1,2}]$. Thus, to preserve the joint distribution while abstracting the state space of a variable, we in general have to introduce additional dependence links between the variable's predecessors and successors (Chang and Fung 1991). But introducing such links would defeat our original purpose for abstraction, namely to trade accuracy for speed. Hence, we generally choose to forego preservation of the joint (1), in return for inferential efficiency.

There is another problem with using (3) to translate the OPN to an APN. For example, consider the operation of abstracting $b$ in our simple 3-node network above. The revised conditional probability for $c$ given the new superstate is:

$$\Pr\left(c \middle| [b_{1,2}]\right) = \frac{\Pr(c|b_1)\Pr(b_1) + \Pr(c|b_2)\Pr(b_2)}{\Pr(b_1) + \Pr(b_2)}. \tag{4}$$

But whereas the conditional probabilities $\Pr(c|b_i)$ are specified as part of the OPN, the marginal probabilities $\Pr(b_i)$ are not. We could compute them from the OPN, but of course this would require evaluating the network at its finest granularity, which is what we are trying to avoid in the first place.

Examining (4), we see that the conditional distribution for $c$ given a superstate is a weighted sum of terms conditional on each of the constituent elementary states. Since deriving valid weights requires unavailable marginals, we instead weight the terms uniformly. In this case, the approximate conditional is

$$\Pr\left(c \middle| \left[b_{1,2}\right]\right) = \frac{1}{2}\Pr\left(c \middle| b_1\right) + \frac{1}{2}\Pr\left(c \middle| b_2\right). \tag{5}$$

We call this approach the *average policy*. If we had some way of approximating the marginals, these approximations could be substituted for the uniform weights. For example, Chang and Fung (1991) suggest that we instead use the relative *conditional* probabilities for the $b_i$ given $b$'s predecessors (in this case, $a$), averaged over the possible values of the predecessors.

### 3.3. AN ITERATIVE ABSTRACTION PROCEDURE

The basic idea behind our abstraction procedure is to start with a very abstract network and iteratively refine the most probable superstates. We generate an initial APN by replacing the state space of each node in the OPN with a single superstate. We then split each superstate as follows:

**Definition 5** To *split* a superstate $\left[s_{i,j}\right]$ we replace it with the pair of states $\left[s_{i,k}\right]$ and $\left[s_{k+1,j}\right]$, where $k = \lfloor (i+j-1)/2 \rfloor$. (The first is an elementary state if $k = i$, and the second is elementary if $j = k+1$.) Conditional probability distributions are translated from the OPN using the average policy.

The strategy for refining the APNs can have a great influence on the performance of the iterative procedure. Our approach is to split, at each iteration, the most probable superstates of each node. This requires an evaluation of the marginal distribution of each node, at each iteration. We continue iterating until either we run out of time, or run out of superstates to split. The procedure is summarized in Figure 2. It can be interrupted at any time, in which case it returns the latest evaluation results.

**procedure** Abstract-Iter(OPN, evidence)

1. Generate an initial APN with one superstate per node.
2. Evaluate the probability distribution for each node given the evidence.
3. If all states for all nodes are elementary, return.
4. Split the most probable superstate in each node.
5. Go to step 2.

**Figure 2**: The iterative abstraction procedure.

Several variations on this procedure are readily conceivable. First, we could at each iteration split only a single superstate, rather than one superstate per node. Second, we could choose different heuristic policies for choosing which superstate to split, for example we could consider the difference between the conditional probabilities among the elementary states. Third, we might adopt alternate policies for weighting the elementary terms in computing conditional probabilities for the APN. Finally, we could include some of the dependencies necessary to preserve the joint distribution during abstraction. We intend to explore these and other variations in further work; the main results we present below have been generated with respect to our baseline procedure.

Note that our iterative procedure contains much redundancy, in recomputing the marginals from scratch for each iteration. We expect that performance could (and should!) be improved significantly by making the evaluation more incremental (D'Ambrosio 1993). To achieve incrementality requires that we modify the internals of the basic evaluation procedure—a step that we have avoided thus far but is undoubtedly necessary for practical use of the procedure.

### 3.4. EMPIRICAL RESULTS

We have run our iterative abstraction procedure on several networks, including that of Figure 1. In each experiment, the intermediate and final probability distributions were compared with the exact solution computed using the finest-grain OPN.

We adopt a standard proper scoring rule to measure the quality of the approximate solutions produced by the abstraction procedure. Let $o_i = \Pr(s = s_i)$ be the marginal probability that the variable of interest takes its $i$th elementary state, according to the OPN. Evaluating the APN yields marginal probabilities of the form $a_{i,j} = \Pr(s_i \le s \le s_j)$ for each superstate $\left[s_{i,j}\right]$ of node $s$ in the APN (elementary states can be treated as a special case, with $i = j$). To compare the two distributions, we interpret the probability of the superstate as a *uniform* distribution over its constituent elementary states. That is, we define

$$a_k = \frac{a_{i,j}}{j-i+1}, \text{ for } i \le k \le j.$$

For a variable with $m$ elementary states, we use the logarithmic scoring rule

$$score(a) = \sum_{i=1}^{m} o_i \log a_i,$$

which is *proper* in that it is maximized when $a_i = o_i$ for all $i$. The measure is well-behaved as long as $a_i > 0$ whenever $o_i > 0$, which holds for our algorithm. Taking the maximum possible score to be $score(o)$, we can measure the relative score for a particular approximation $a$:

$$relscore(a) = \frac{score(o)}{score(a)},$$

Since $score(a) \le score(\bullet) \le 0$, the relative score lies in the interval [0,1], with higher values corresponding to better approximations. We can measure the overall relative score for a network by averaging the relative scores for each node.

### 3.4.1. Test Case 1: Simple Commuter Model

Our first test case is based on the simple commuter model of Figure 1. The goal was to compute the marginal distribution of every node, given as evidence a particular value for *LH*. The following graph presents the relative scores for three versions of the network. For each, we plot the average relative score as a function of time, where each time point corresponds to a distinct iteration of our anytime abstraction procedure. The first point in each series (off the scale at 0.82 for test3) represents the initial APN, with one superstate per node. In this initial situation the approximation is simply the uniform distribution, which serves as a baseline for our *relscore* measure of fit.

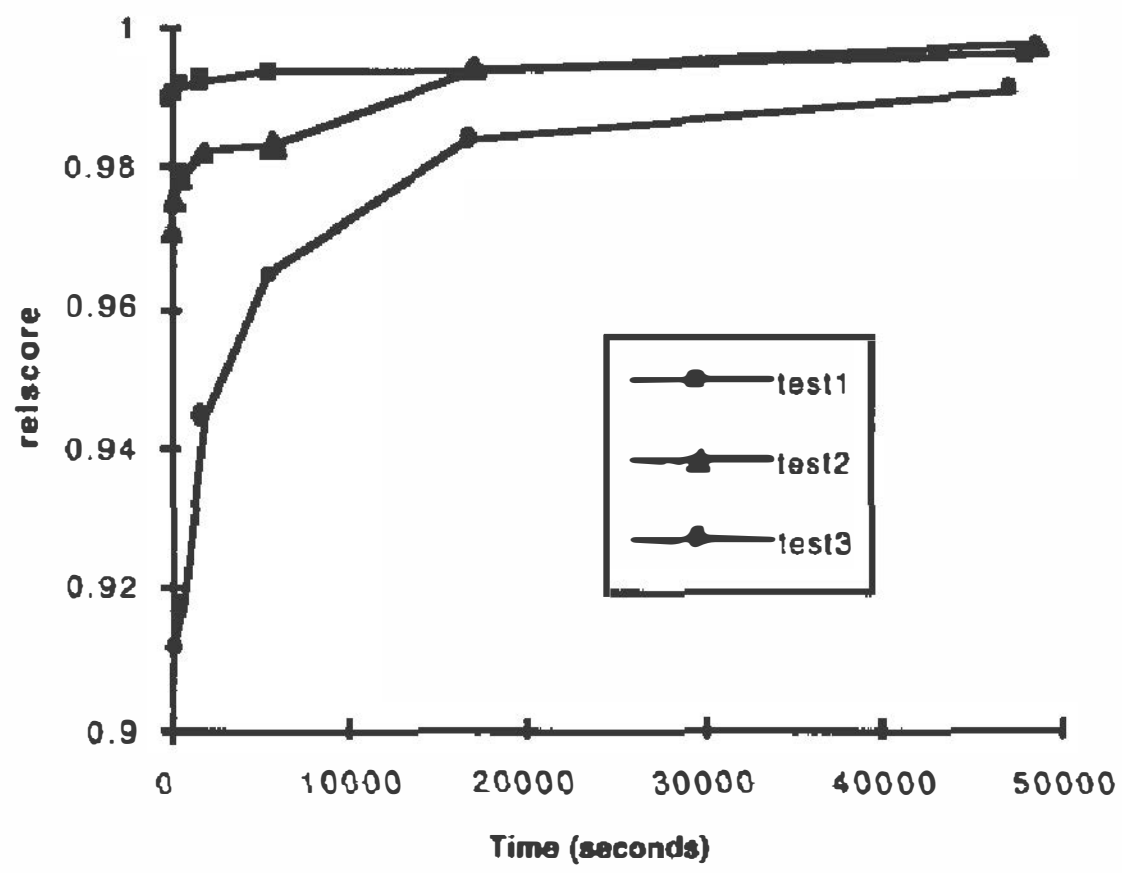

**Figure 3**: Iterative abstraction results.

As we can see in the graph, the approximation improves monotonically, approaching the exact distribution in the limit (when the refinement reaches the elementary states, the result is exact). In these examples, the OPN had eight states per node, and the graph presents values for iterations one through seven. At each iteration, we evaluate the probability of each node given the evidence, using Shachter's graph-reduction method (Shachter 1988).[3] In iteration *k*, each node has *k* states. Note that the time per iteration increases substantially (exponentially) as we proceed. Since the proportion of time spent on early iterations becomes negligible, there is relatively little advantage to estimating the maximum granularity solvable in a given time and proceeding right to that level. Moreover, the earlier iterations determine the refinement pattern (i.e., which superstates to split); uniform refinement at a preidentified granularity would not be as accurate.

The difference between the tests lies in the probabilistic relationships quantifying the network. In test1 we generated the probabilistic parameters by sampling from a uniform distribution. This we considered to be a favorable case for our approximation method, as both the average policy and our interpretation of the probability of superstates make use of uniformity. Indeed, in this model the relative score is high (0.99) even before any refinement! Even starting from this level, however, the approximation improves smoothly with refinement. Test2 is similar, except that the probabilistic parameters were generated from a skewed distribution. The initial fit was somewhat worse (0.97), but after five iterations was superior to the test1 case. Finally, test3 had the most skewed parameters, with many relationships modeled deterministically. The improvement with refinement in this case was more substantial, reaching a much better fit in just a few iterations.

### 3.4.2. Test Case 2: Multistage Traffic Model

Our second set of experiments reproduce the same qualitative behavior on a different traffic model, using various numbers of stages and fidelities, implementing the abstraction procedure with a different network evaluation algorithm, in a different computational environment.

The second traffic model is a multistage network based on the fundamental equation of traffic flow on uncongested networks, $q = uk$ (flow = speed times concentration). We represent the arrival time at locations A,B,C,... by nodes TA,TB,TC,..., each dependent on the arrival time at the previous node and the speed traveled in the interim. This speed is dependent probabilistically on concentration and flow, which are in turn time-dependent and uncertain. The first two stages of such a model are depicted in Figure 4.

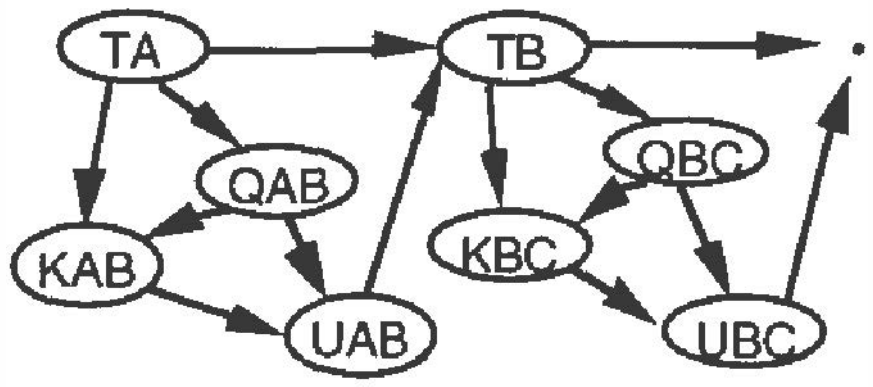

**Figure 4**: Multistage traffic model.

We have applied our iterative abstraction procedure to various versions of this model. The results presented below were derived from five-stage networks (20 nodes), with 24 states per node. We assumed a prediction task, with node TA the only evidence. Figure 5 depicts the average relscore as a function of time for five different instantiations of the network.[4] The cases differ in the sharpness of the conditional distributions, as determined by a global "sd" (standard deviation) parameter.

[3]Implemented in IDEAL, running on a Macintosh Quadra 650 with 20MB RAM allocated to Lisp. The exact values were calculated using Jensen's method; state spaces of cardinality 8 were beyond feasibility using the graph-reduction algorithm (even without reversals). We did not use Jensen's algorithm for our iterative procedure because in this computational setup it would incur an unacceptable overhead of regenerating the join tree at every stage. This problem is mitigated in our next test case, below. In future work, we intend to make the entire process incremental, eviscerating our basic evaluation algorithms as necessary.

[4]Implemented in HUGIN™, running on a Sun SparcStation 2. Both the exact OPN values and the iterative belief evaluations were calculated using Jensen's method. We computed a new join

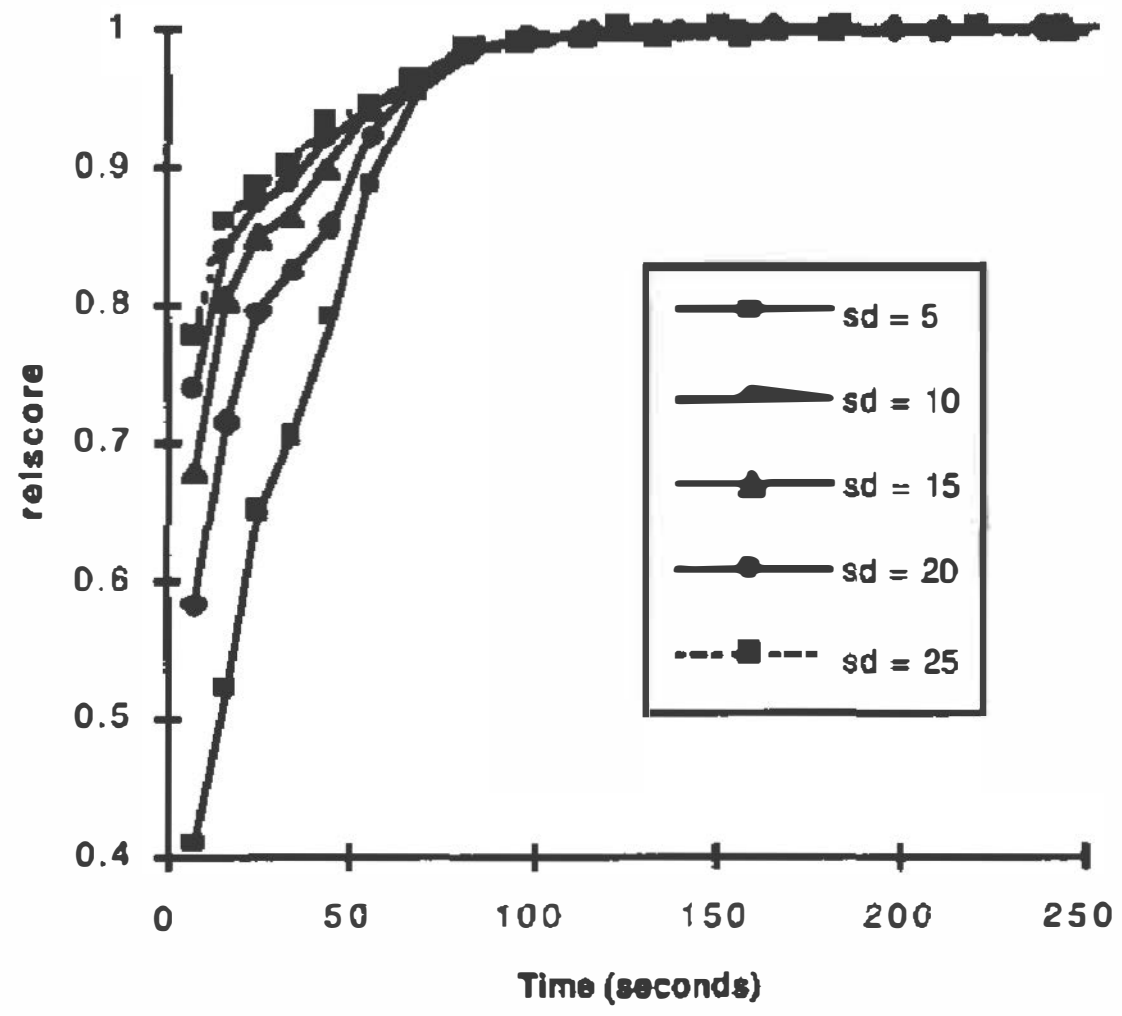

**Figure 5**: Abstract-Iter on the 5-stage traffic model.

As expected, the sharper distributions (lower sd) are initially approximated much worse by the coarse-grained models. However, within a few iterations, the approximations for all cases are quite good. For a range of smaller sd values, we found that there is a granularity point where these are approximated *better*, since more of the probability mass is accounted for by a relatively smaller number of states.

The main point, however, is that the curves consistently exhibit the pattern we look for in an anytime algorithm: rapid initial improvement, converging on the exact answer. Evaluation of these models at full granularity (24 states per node) takes on the order of 125 seconds, which was roughly equivalent to ten abstraction iterations in our experiments. The approximations have a relscore of 0.99 at this point, with the extra advantage of having produced useful approximations even earlier.

We have observed similar behavior for other combinations of the basic model parameters: number of stages, number of states, and sharpness of distributions. Likewise, in limited tests, instantiating downstream nodes (evidential as opposed to causal reasoning) did not appear to affect the observed behavior.

### 3.4.3. Algorithmic Variations

We have begun a limited investigation of some of the variations on the basic Abstract-Iter algorithm mentioned above. Specifically, we consider (1) an alternate method for deriving the conditional probabilities in the APN, and (2) alternate ways to choose which state to refine next.

tree at each iteration, although we needed to triangulate the graph only once. Times reported include this compilation overhead but do *not* include file I/O required in interfacing between our iterative abstraction code and HUGIN.

For simplicity, we examined the most trivial example, that of the three-node network

$$a \rightarrow b \rightarrow c.$$

We suppose that $a$ and $c$ are binary, and that $b$ has $n$ elementary states. We take $n = 64$ in the studies below. Conditional probabilities for $c$ given $b$ and $b$ given $a$ are assigned randomly. The query is for the marginal $\Pr(c)$.

In our first test, we compared the average policy (5) for assigning the conditional probability of $c$ given a superstate of $b$ with that recommended by Chang and Fung (1991). In this "CF" policy, rather than weight the elementary states $b_i$ uniformly, we weight them according to their conditional probabilities given $b$'s predecessor, $a$. We would expect the CF policy to perform somewhat better, at the cost of applying more information. This is confirmed by the charts of Figure 5. Each chart represents 100 randomly parametrized networks, with the "rel error" axis measuring the average percentage disparity between $\Pr(c)$ in the APN and the OPN.

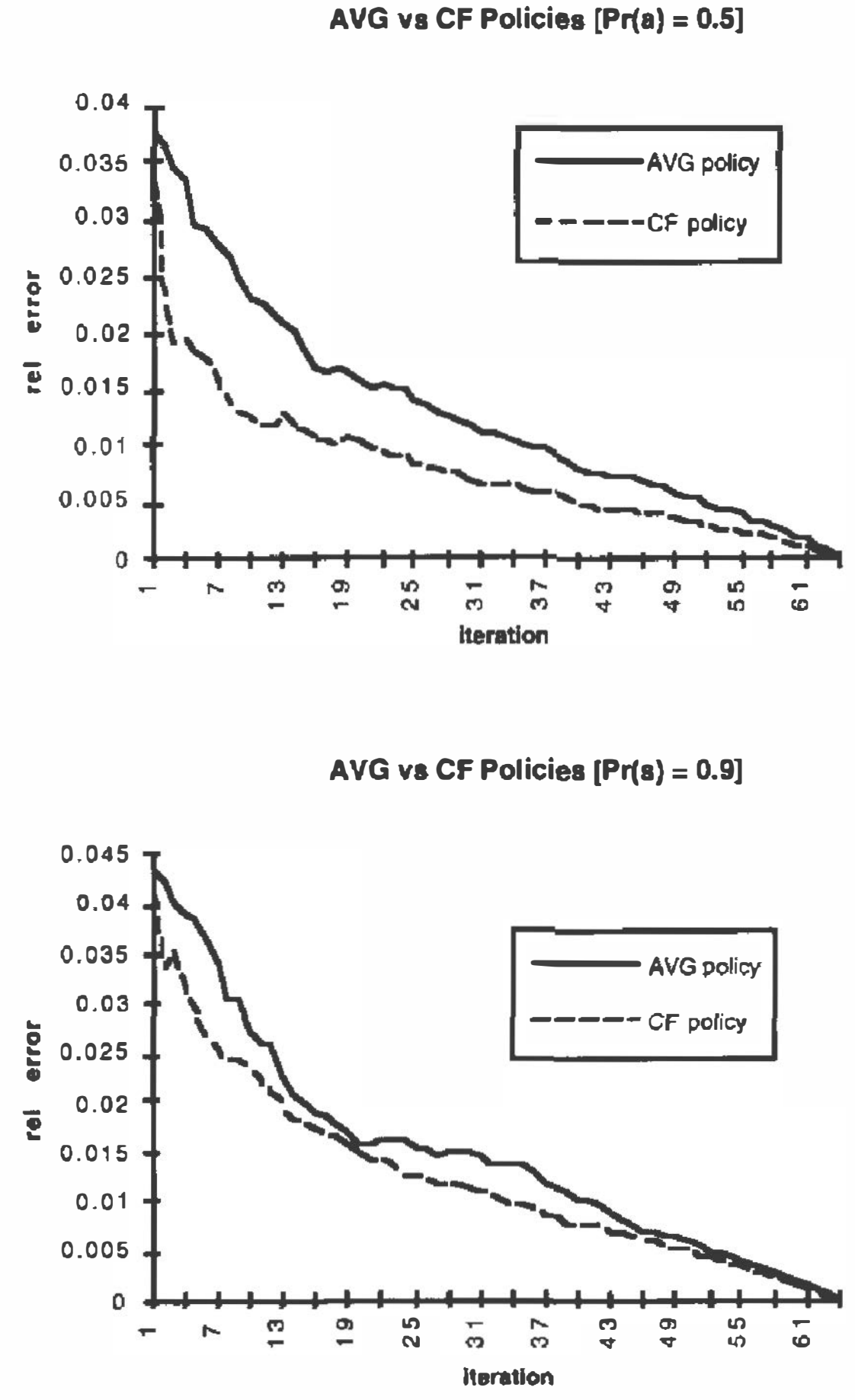

**Figure 5:** Comparison of average and CF policies.

The first chart represents the case where $\Pr(a) = 0.5$, which is the most favorable case for the CF method, as it produces the exact conditional probabilities. When the predecessors of $b$ are more skewed, as in the second chart, the advantage of taking into account $b$'s distribution is not as great.

We have also begun to explore alternate criteria for choosing which node or state to refine next. Although the results are far from definitive, preliminary experience suggests that the skew among the conditional probabilities of a node's substates should be taken into account (as a complementary factor to overall likelihood).

# 4. MODEL STRUCTURE ABSTRACTION

Although the focus of this paper is on state-space abstraction, we believe that structural abstraction is another important technique for anytime evaluation of probabilistic networks. For example, it would be unwise to generate the probabilistic network corresponding to a complete road map of the United States in order to compute the distribution of travel times from the University of Michigan AI Laboratory to the Ann Arbor Public Library. In this case we would primarily focus on bounding the spatial scope of the model. We would also avoid using the entire map even when driving from Ann Arbor to Seattle, although in this case more than spatial scoping is necessary. In this situation, we would apply abstraction to the road network, in order to focus on the major highways and ignore the secondary and tertiary roads. However, we may need to restore this detail at a later time, for example when we are looking to stop for lunch near Omaha.

An anytime procedure for structural abstraction would work in a manner similar to our procedure for state-space abstraction. For example, suppose our task is to predict the travel time from $A$ to $A'$ in the road network of Figure 6. Let lines represent roads, with the thickness of lines representing the traffic capacities of road segments. A probabilistic network for predicting the travel time from $A$ to $A'$ is shown in Figure 7. In the network, nodes named $TX$ represent the time that the driver arrives at location $X$, and nodes named by a pair of locations represent the time needed to travel from one location to the other.

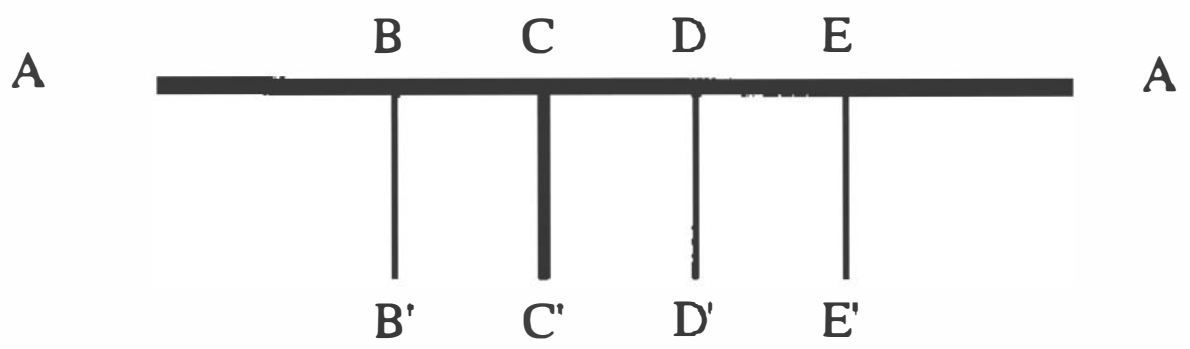

**Figure 6**: A simple road network.

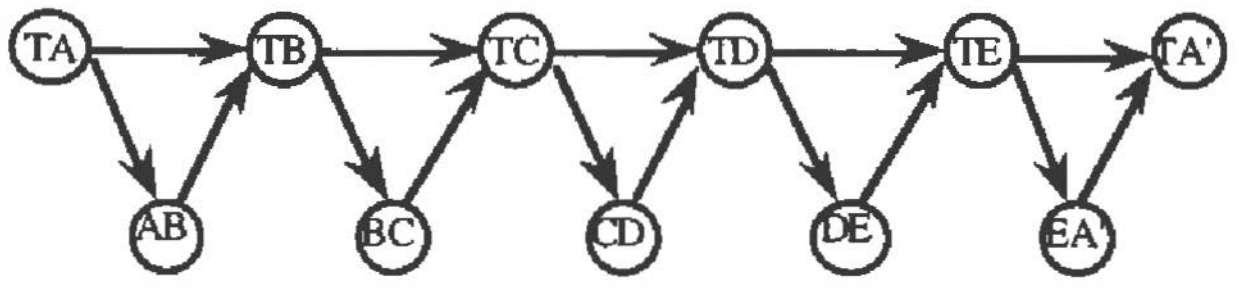

**Figure 7**: Detailed probabilistic network for travel-time prediction.

If the detailed probabilistic network is too complicated to solve, we can start with a more abstract model of the problem. For example, one abstract model (Figure 8(a)) might directly specify the approximate distribution of travel times between $A$ and $A'$. If we have some more time, we might entertain further details by breaking the road into smaller segments. The intersection at $C$ might be particularly significant, since there is a moderately wide road connecting to $AA'$ at that point. We can refine the network of Figure 8(a) structurally, resulting in the network of 8(b). By iteratively considering smaller road segments, we define an anytime approximation algorithm based on structural, as opposed to state-space, abstraction.

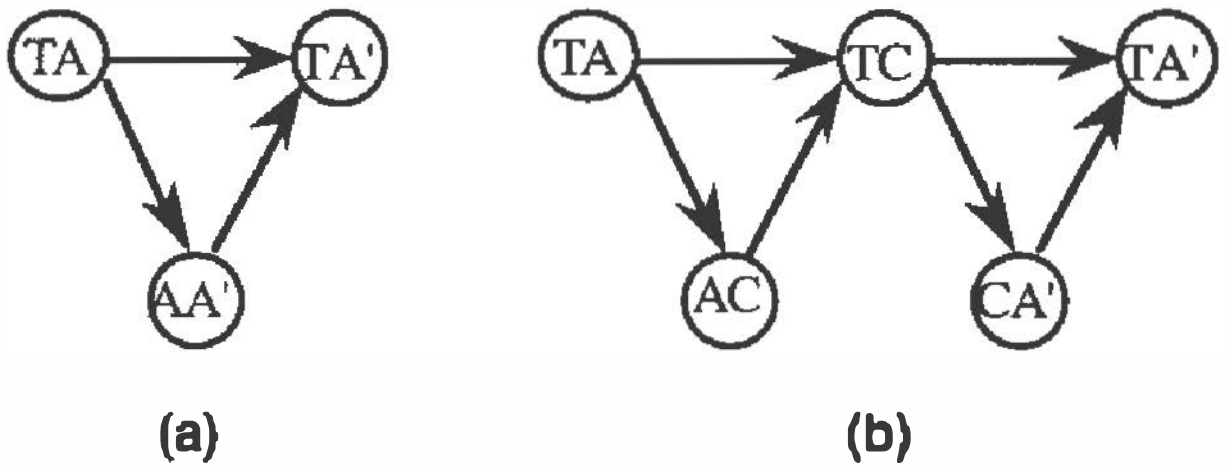

**Figure 8**: Structural abstraction: (a) simplest model, (b) slightly refined model.

Structural and state-space abstraction are complementary approaches to probabilistic-network approximation. They are also related, as abstracting the state space of a node to a single superstate is tantamount to ignoring the node. (Although doing this without modifying the dependence structure can alter the joint distribution, as noted above.)

# 5. MORE RELATED WORK

Although we are aware of no previous investigations specifically employing state-space abstraction for anytime query-processing in probabilistic networks, the idea is clearly "in the air", related to many other research efforts.

We have enumerated some other anytime approaches to probabilistic inference in the introductory section above. Other approximation schemes for probabilistic networks have also been proposed, involving deletion of nodes, links, or states (Jensen and Andersen 1990; Kjærulff 1993; Sarkar 1993). The uncertain reasoning literature has also seen some alternate approaches to abstraction in probabilistic networks (Poh et al. 1994) and probabilistic reasoning (Horvitz and Klein 1993). Finally, there has been some investigation of the general problem of reasoning about the quality of approximate models (Laskey and Lehner 1994). We are currently exploring the relations among all these approaches and our own work.

# 6. CONCLUSION

The foregoing experience suggests that iterative refinement is a viable anytime approach to approximate evaluation of probabilistic networks. However, there are several ways in which the current procedure could be improved. In particular, there is no good reason to refine all the nodes in lockstep; refining some nodes will clearly have more benefit than others. Future work will investigate this and several other options, and compare the most effective methods we find with alternate approximation strategies.

We also lack at this time precise analytical models relating the quality of approximation with degree of refinement. Bounding the error incurred by coarsening variables is difficult, as in the worst case (at least locally), it can be almost arbitrarily bad. Nevertheless, we intend also to attempt to characterize as well as possible the improvement that may be expected via refinement, and the potential errors induced by alternate abstraction policies.

Despite these gaps in our understanding, it seems clear that state-space granularity is one of the important control knobs in the design of real-time probabilistic reasoners. Progressive improvement via iterative refinement is one way to twiddle this knob, and is a particularly simple twiddle to embed in an anytime evaluation algorithm.

## Acknowledgments

We thank the anonymous referees for prompting us to run more experiments. This work was supported in part by Grant F49620-94-1-0027 from the Air Force Office of Scientific Research.

## References

Breese, J. S. (1992). Construction of belief and decision networks. *Computational Intelligence* **8**(4): 624-647.

Breese, J. S., R. P. Goldman, and M. P. Wellman (1994). Special Section on Knowledge-Based Construction of Probabilistic and Decision Models. *IEEE Transactions on Systems, Man, and Cybernetics* **24**(11).

Breese, J. S., and E. J. Horvitz (1991). Ideal reformulation of belief networks. *Uncertainty in Artificial Intelligence 6* Ed. P. P. Bonissone et al. North-Holland. 129-143.

Chang, K.-C., and R. Fung (1991). Refinement and coarsening of Bayesian networks. *Uncertainty in Artificial Intelligence 6* Ed. P. P. Bonissone et al. North-Holland.

Charniak, E. (1991). Bayesian networks without tears. *AI Magazine* **12**(4): 50-63.

Cooper, G. F. (1990). The computational complexity of probabilistic inference using Bayesian belief networks. *Artificial Intelligence* **42**: 393-405.

D'Ambrosio, B. (1993). Incremental probabilistic inference. *Proceedings of the Ninth Conference on Uncertainty in Artificial Intelligence*, Washington, DC, Morgan Kaufmann.

Dagum, P., and M. Luby (1993). Approximating probabilistic inference in Bayesian belief networks is NP-hard. *Artificial Intelligence* **60**: 141-153.

Ezawa, K. J. (1986). *Efficient Evaluation of Influence Diagrams*. PhD Thesis, Stanford University.

Goldman, R. P., and E. Charniak (1993). A language for construction of belief networks. *IEEE Transactions on Pattern Analysis and Machine Intelligence* **15**: 196-208.

Horvitz, E. J., and A. C. Klein (1993). Utility-based abstraction and categorization. *Proceedings of the Ninth Conference on Uncertainty in Artificial Intelligence*, Washington, DC, Morgan Kaufmann.

Horvitz, E. J., H. J. Suermondt, and G. F. Cooper (1989). Bounded conditioning: Flexible inference for decisions under scarce resources. *Proceedings of the Fifth Workshop on Uncertainty in Artificial Intelligence*, Windsor, ON, Association for Uncertainty in AI.

Jensen, F., and S. K. Andersen (1990). Approximations in Bayesian belief universes for knowledge-based systems. *Proceedings of the Sixth Conference on Uncertainty in Artificial Intelligence*, Cambridge, MA.

Jensen, F. V., K. G. Olesen, and S. K. Andersen (1990). An algebra of Bayesian belief universes for knowledge-based systems. *Networks* **20**: 637-660.

Kjærulff, U. (1993). Approximation of Bayesian networks through edge removals. Institute for Electronic Systems, Aalborg University.

Laskey, K. B., and P. E. Lehner (1994). Metareasoning and the problem of small worlds. *IEEE Transactions on Systems, Man, and Cybernetics* **24**(11).

Neapolitan, R. E. (1990). *Probabilistic Reasoning in Expert Systems: Theory and Algorithms.* Wiley & Sons.

Nicholson, A. E., and J. M. Brady (1994). Dynamic belief networks for discrete monitoring. *IEEE Transactions on Systems, Man, and Cybernetics* **24**(11).

Pearl, J. (1988). *Probabilistic Reasoning in Intelligent Systems: Networks of Plausible Inference.* San Mateo, CA, Morgan Kaufmann.

Poh, K. L., M. R. Fehling, and E. J. Horvitz (1994). Dynamic construction and refinement of utility-based categorization models. *IEEE Transactions on Systems, Man, and Cybernetics* **24**(11).

Provan, G. M. (1994). Tradeoffs in knowledge-based construction of probabilistic models. *IEEE Transactions on Systems, Man, and Cybernetics* **24**(11).

Saffiotti, A., and E. Umkehrer (1994). Inference-driven construction of valuation systems from first-order clauses. *IEEE Transactions on Systems, Man, and Cybernetics* **24**.

Sarkar, S. (1993). Using tree-decomposable structures to approximate belief networks. *Proceedings of the Ninth Conference on Uncertainty in Artificial Intelligence*, Washington, DC, Morgan Kaufmann.

Shachter, R. D. (1988). Probabilistic inference and influence diagrams. *Operations Research* **36**: 589-604.

Srinivas, S., and J. Breese (1990). IDEAL: A software package for analysis of influence diagrams. *Proceedings of the Sixth Conference on Uncertainty in Artificial Intelligence*, Cambridge, MA, Assoc. Uncertainty in AI.

Wellman, M. P., J. S. Breese, and R. P. Goldman (1992). From knowledge bases to decision models. *Knowledge Engineering Review* **7**(1): 35-53.